\documentclass[runningheads]{llncs}
\usepackage{graphicx}
\usepackage{pdfpages}
\usepackage{algpseudocode}
\usepackage{algorithm}
\usepackage{multicol}
\usepackage{multirow}
\usepackage{bm}
\usepackage{bibentry}
\usepackage{listings}
\usepackage{amssymb}
\usepackage{amsmath}
\usepackage{pifont}
\usepackage{cite}
\usepackage{xcolor}
\usepackage{color, colortbl}
\usepackage{orcidlink}
\usepackage{xurl}
\newcommand{\repeatthanks}{\textsuperscript{\thefootnote}}
\newcommand{\etal}{\textit{et al. }}

\definecolor{lightgray}{gray}{0.9}

\begin{document}
\title{Cluster-based Video Summarization with Temporal Context Awareness}
%


\author{
    Hai-Dang Huynh-Lam
    \inst{1, 2}
    \thanks{Both authors contributed equally to this research.}
\and
    Ngoc-Phuong Ho-Thi
    \inst{1, 2}
    \repeatthanks
\and
    Minh-Triet Tran
    \inst{1, 2}
    \orcidlink{0000-0003-3046-3041}
\and
    Trung-Nghia Le
    \inst{1, 2}
    \orcidlink{0000-0002-7363-2610}
}

\authorrunning{Huynh and Ho \etal}

\institute{
University of Science, VNU-HCM, Vietnam
\and
    Vietnam National University, Ho Chi Minh, Vietnam
}

\maketitle              

\begin{abstract}
    In this paper, we present TAC-SUM, a novel and efficient training-free approach for video summarization that addresses the limitations of existing cluster-based models by incorporating temporal context. Our method partitions the input video into temporally consecutive segments with clustering information, enabling the injection of temporal awareness into the clustering process, setting it apart from prior cluster-based summarization methods. The resulting temporal-aware clusters are then utilized to compute the final summary, using simple rules for keyframe selection and frame importance scoring. Experimental results on the SumMe dataset demonstrate the effectiveness of our proposed approach, outperforming existing unsupervised methods and achieving comparable performance to state-of-the-art supervised summarization techniques. Our source code is available for reference at \url{https://github.com/hcmus-thesis-gulu/TAC-SUM}.
    
    
\end{abstract}

\section{Introduction}
\label{sec:intro}

    Video summarization is a crucial research area that aims to generate concise and informative summaries of videos, capturing their temporal and semantic aspects while preserving essential content. This task poses several challenges, including identifying important frames or shots, detecting significant events, and maintaining overall coherence. Video summarization finds applications in diverse fields, enhancing video browsing, retrieval, and user experience \cite{Apostolidis2021Video}.
    
    The current state-of-the-art methods in summarizing videos are SMN \cite{wang2019stacked} and PGL-SUM \cite{apostolidis2021pglsum}. 
    SMN stacks LSTM and memory layers hierarchically to capture long-term temporal context and estimate frame importance based on this information. Its training, however, relies on LSTMs and is not fully parallelizable.
    PGL-SUM uses self-attention mechanisms to estimate the importance and dependencies of video frames. It combines global and local multi-head attention with positional encoding to create concise and representative video summaries. 
    Both SMN and PGL-SUM heavily rely on human-generated summaries as ground truth, introducing biases and inconsistencies during training.

    To eliminate the need for labeled data required by supervised approaches, unsupervised algorithms have been explored, such as Generative Adversarial Networks \cite{apostolidis2020unsupervised} and Reinforcement Learning \cite{zhou2018deep}. While achieving remarkable results without annotations, their performance gains have been minor compared to supervised methods, and the computational requirements can be high with GPU usage.

    A line of research focusing on the use of clustering algorithms for video summarization has been pioneered by De \etal~\cite{de2011vsumm} and followed by Mahmoud \etal~\cite{mahmoud2013unsupervised} to create interpretable summaries without labels and training. Such methods demonstrate acceptable performance in low-resource environments, but their effectiveness has yet to be competitive with learnable approaches.
    
    In this paper, we propose a training-free approach called Temporal-Aware Cluster-based SUMmarization (TAC-SUM) to address the challenges encountered by previous studies. This method leverages temporal relations between frames inside a video to convert clusters of frames into temporally aware segments. Specifically, frame similarities available from these clusters are used to divide the video into non-overlapping and consecutive segments. The proposed algorithm then applies simple and naive rules to select keyframes from these segments as well as assign importance scores to each frame based on its segment's information. Our approach is expected to outperform existing cluster-based methods by injecting temporal awareness after the clustering step. It eliminates the need for expensive annotation, increases efficiency, and offers high interpretability due to its visualizability and transparent rules. An important distinction from some previous unsupervised studies is that TAC-SUM currently relies on naive rules, leaving ample room for future improvement, including the integration of learnable components, which have been successful in learning-free algorithms \cite{mahmoud2013unsupervised}.

    We conduct quantitative and qualitative experiments on the SumMe dataset \cite{SumMe} to evaluate our method's performance in video summarization. The quantitative experiment shows that our approach significantly outperforms existing unsupervised methods and is comparable to current state-of-the-art supervised algorithms. The qualitative study demonstrates that our approach produces effective visual summaries and exhibits high interpretability with the use of naive rules. 
    
    The main contributions presented in the paper are as follows:
    \begin{itemize}
        \item We introduce the integration of temporal context into the clustering mechanism for video summarization, addressing the shortcomings of traditional cluster-based methods.
        \item We propose a novel architecture that effectively embeds temporal context into the clustering step, leading to improved video summarization results.
        \item Our approach demonstrates superior performance compared to existing cluster-based methods and remains competitive with state-of-the-art deep learning summarization approaches.
    \end{itemize}

    
\section{Related Work}
\label{sec:related}
    Video summarization techniques can be broadly classified into two categories: supervised methods and unsupervised methods. While supervised methods demonstrate superior performance in domain-specific applications, they rely heavily on labeled data, making them less practical for general video summarization tasks where labeled data may be scarce or costly to obtain. As a result, unsupervised methods remain popular for their versatility and ability to generate summaries without the need for labeled data. Within unsupervised approaches, clustering algorithms have emerged as a popular choice.

    Cluster-based video summarization methods utilize the concept of grouping similar frames or shots into clusters and selecting representative keyframes from each cluster to form the final summary. These approaches have shown promise in generating meaningful summaries, as they can capture content diversity and reduce redundancy effectively.
    Prior works have explored various clustering techniques for video summarization. Mundu \etal \cite{mundur2006keyframe} employed Delaunay triangulation clustering using color feature space, but high computational overhead limited its practicality. De \etal \cite{de2011vsumm} utilized K-means clustering with hue histogram representation for keyframe extraction. Shroff \etal \cite{shroff2010video} introduced a modified version of K-means that considers inter-cluster center variance and intra-cluster distance for improved representativeness and diversity. Asadi \etal \cite{asadi2012video} applied fuzzy C-means clustering with color component histograms. Mahmoud \etal \cite{mahmoud2013vscan} used DBSCAN clustering with Bhattacharya distance as a similarity metric within the VSCAN algorithm.
    Cluster-based methods offer simplicity and interpretability, often relying on distance metrics like Euclidean or cosine similarity to group similar frames. Their computational efficiency allows for scalability to large video datasets. However, traditional cluster-based approaches have limitations. Notably, they may overlook temporal coherence, leading to fragmented and incoherent summaries. Additionally, handling complex video content with multiple events or dynamic scenes can pose challenges, as these methods primarily rely on visual similarity for clustering.

    With the rise of deep learning, video summarization has seen significant advancements. In supervised approaches, temporal coherence is addressed by modeling variable-range temporal dependencies among frames and learning their importance based on ground-truth annotations. This has been achieved using various architectures, such as LSTM-based key-frame selectors \cite{zhang2016lstm, zhao2017hierarchical, zhao2018hsa, zhao2020tth}, Fully Convolutional Sequence Networks \cite{rochan2018sequence}, and attention-based architectures \cite{fajtl2019summarizing, li2021exploring, liu2019learning}.

    However, achieving temporal coherence in unsupervised learning poses challenges. One promising direction is the utilization of Generative Adversarial Networks (GANs). Mahasseni \etal \cite{mahasseni2017unsupervised} combined an LSTM-based key-frame selector, a Variational Auto-Encoder (VAE), and a trainable Discriminator in an adversarial learning framework to reconstruct the original video from the summary. Other works extended this core VAE-GAN architecture with tailored attention mechanisms to capture frame dependencies at various temporal granularities during keyframe selection \cite{jung2019discriminative, jung2020global, he2019unsupervised}. These methods focus on important temporal regions and model long-range dependencies in the video sequence.
    Although GAN-based models have shown promise in generating coherent summaries, they face challenges of unstable training and limited evaluation criteria. 
    
    The proposed method leverages cluster-based models by utilizing visual representations generated by unsupervised deep learning approaches such as DINO \cite{dino}. Addressing the problem of temporal coherence, our developed TAC-SUM introduces the temporal context into the process. This integration of temporal context enhances the summarization performance, as demonstrated by experimental results.

\section{Proposed Approach}
\label{sec:method}

    

\begin{figure}[!t]
        \centering
        \includegraphics[width=0.5\paperwidth]{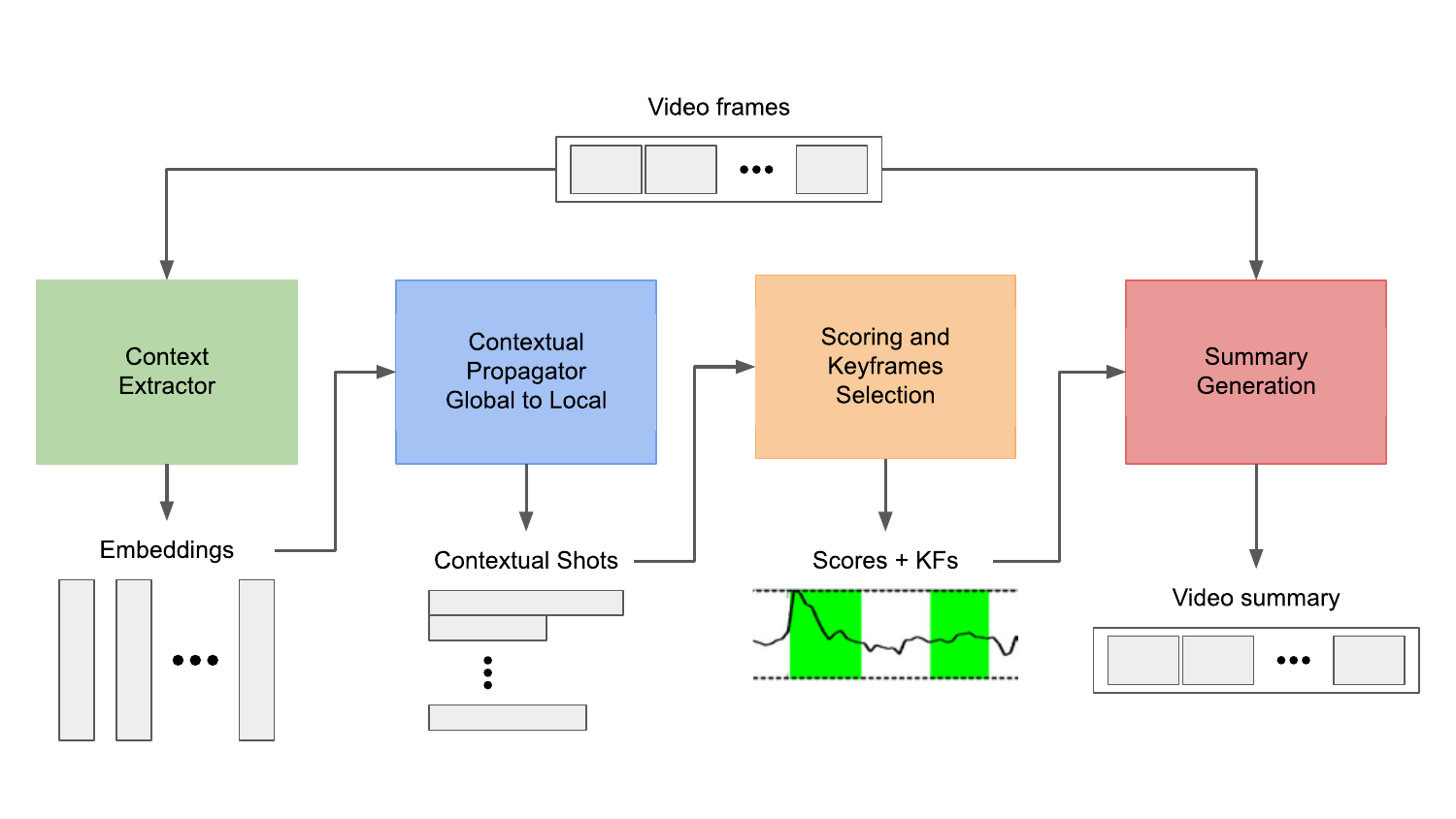}
        \caption{Pipeline of the proposed approach showcasing four modules and information flow across main stages.}
        \label{fig:method-pipeline}
    \end{figure}

    Our approach selects an ordered subset $\mathbf{S} = \{I_{t_1}, I_{t_2}, \ldots, I_{t_L}\}$ of $L$ frames from a video $\mathbf{I} = \{I_1, I_2, \ldots, I_T\}$, where $T$ is the total number of frames and the summarized subset $\mathbf{S}$ is obtained by selecting frames indexed at $t_i$ positions. The timestamp vector $\mathbf{t}$ comprises such positions $\{t_1, t_2, \ldots, t_L\}$.
    In Figure \ref{fig:method-pipeline}, we illustrate the four stages of our method as distinct modules. Each stage comprises several steps tailored to the specific role and algorithm implemented. We provide a detailed explanation of each stage in the remaining text of this section. In addition, Section \ref{subsec:method-imple} is dedicated to clarifying several technical details related to the implementation of our approach.
    
    \subsection{Generating Contextual Embeddings}
    \label{subsec:method-generating}
        This stage extracts the context of an input video $\mathbf{I}$ from its frames $I_t$. It involves two steps: sampling the video and constructing embeddings for each sampled frame.
        
        \paragraph{\textbf{Sampling Step}}
        \label{para:method-generating-sampling}
            To reduce computational complexity, we employ a sampling technique to extract frames from $\mathbf{I}$ into a sequence of samples $\mathbf{\hat{I}}$. The frame rate of $\mathbf{\hat{I}}$ is matched to a pre-specified frame rate $R$. This method ensures representativeness and serves as normalization for different inputs. The sampling process involves dividing the original frames within a one-second period into equal-length snippets and selecting the middle frame of each snippet as the final sample.
        
\begin{figure}[!t]
                \centering
                \includegraphics[width=0.5\paperwidth]{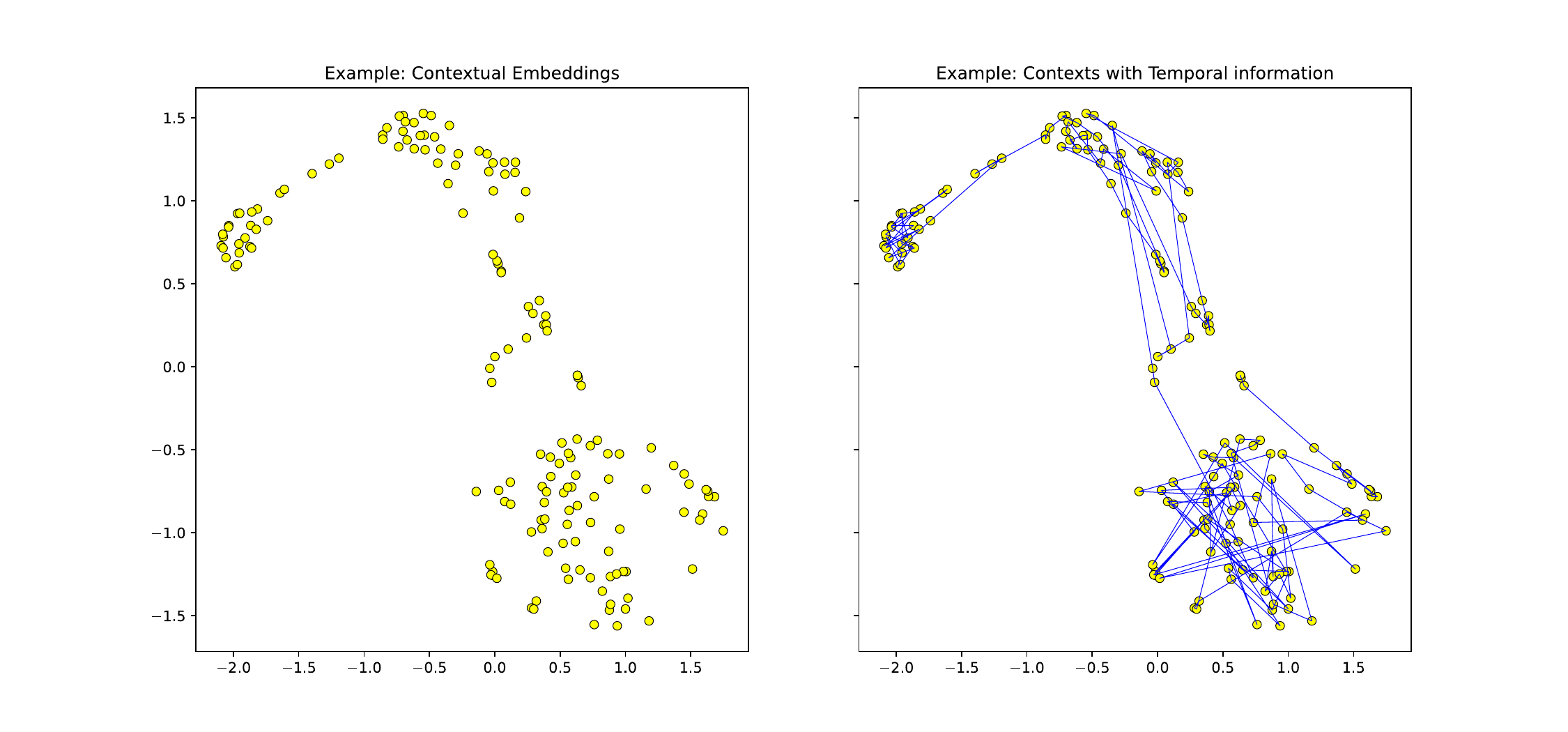}
                \caption{Visual illustration of contextual information.}
                \label{fig:method-context}
            \end{figure}

        \paragraph{\textbf{Embedding Step}}
        \label{para:method-generating-embeddings}
            For each sampled frame $\hat{I}_i$, we utilize a pre-trained model to extract its visual embedding $\mathbf{e}_i$. The pre-trained model is denoted as a function $g: \mathbb{R}^{W \times H \times C} \longrightarrow \mathbb{R}^{D}$ that converts $\hat{I}_i$ into an embedding vector of size $D$. All embeddings are concatenated to form the contextual embedding of the sampled video $\mathbf{E} = \{\mathbf{e}_1, \mathbf{e}_2, \ldots, \mathbf{e}_{\hat{T}}\}$. Figure \ref{fig:method-context} gives two examples of contextual embeddings.
            
    \subsection{From Global Context to Local Semantics}
    \label{subsec:method-global-local}
        This stage distills global information from the contextual embedding $\mathbf{E}$ into finer, local levels. Our method comprises two steps: using traditional clustering to propagate contextual information into partition-level clusters, and further distilling partition-level information into sample-level.
        
\begin{figure}[!t]
                \centering
                \includegraphics[width=\textwidth]{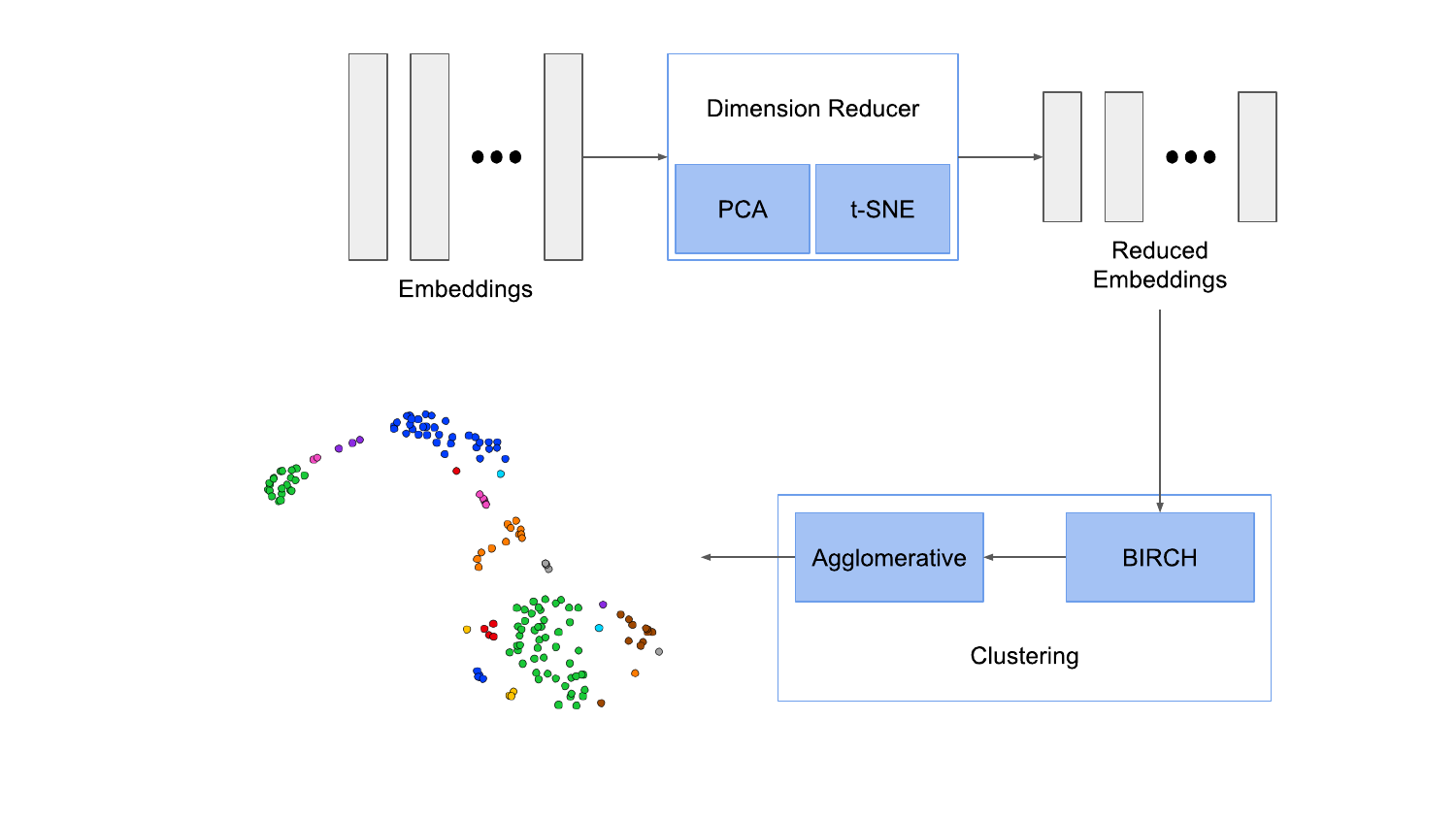}
                \caption{Overall pipeline for the Contextual Clustering step.}
                \label{fig:method-clustering}
            \end{figure}

        \paragraph{\textbf{Contextual Clustering}}
            Clustering the contextual embeddings $\mathbf{E}$ captures global and local relationships between visual elements in the video. We first reduce the dimension of $\mathbf{E}$ to a reduced embedding $\mathbf{\hat{E}}$. A coarse-to-fine clustering approach is then applied to divide the sampled frames into $K$ clusters, creating a label vector $\mathbf{c} \in \mathbf{N}^{\hat{T}}$. More details can be found in Figure \ref{fig:method-clustering}. Starting with the contextual embedding $\mathbf{E} \in \mathbb{R}^{\hat{T} \times D}$, a reduced embedding $\mathbf{\hat{E}} \in \mathbb{R}^{\hat{T} \times \hat{D}}$ is computed using PCA and t-SNE. A traditional clustering method called BIRCH algorithm \cite{birch} is applied to compute coarse clusters of sampled frames, creating a sample-level notation for coarse clusters $\mathbf{\hat{c}} = \{{\hat{c}}_1, {\hat{c}}_2, \ldots, {\hat{c}}_{\hat{T}}\}$. Then, a hierarchical clustering algorithm is employed to combine coarse clusters into finer clusters with the number of eventual clusters is pre-determined based on a sigmoidal function and a maximum threshold. The fine cluster is formed as the union of at least one coarse cluster. Clusters are progressively merged based on affinity between them. This approach achieves a hierarchical clustering that effectively propagates information from the global level $\mathbf{\hat{E}}$ to the local level $\mathbf{c}$, enabling us to extract semantically meaningful clusters.
            
        \paragraph{\textbf{Semantic Partitioning}}

            Following the contextual clustering step, each sampled frame $\hat{I}_i$ is assigned a label ${c}_i$ corresponding to its cluster index. An outlier elimination removes possible outliers and a refinement step consolidates smaller partitions into larger ones with a threshold $\epsilon$. A smoothing operation is applied to labels by assigning the final label ${\hat{c}}_i$ of each frame by taking a majority vote among its consecutive neighboring frames. Once frames have been assigned their final labels $\mathcal{C}$, they are partitioned into sections $\mathcal{P}$ based on these labels.
            The semantic partitioning $\mathcal{P} = \left\{\mathcal{P}_1, \mathcal{P}_2, \ldots, \mathcal{P}_{\hat{N}}\right\}$ obtained from the above process contains $\hat{N}$ sections which are then progressively refined with length condition. Algorithm for this refinement is delineated as follows with a parameter $\epsilon$ denoting the minimum partition's length allowed in the result. Initially, the number of partitions $N$ is set to $\hat{N}$. Subsequently, while the minimum length of the partitions is less than $\epsilon$, the index of the shortest partition $\hat{i}$ is determined.
            The left and right sides of partition $\hat{i}$ are merged with their respective neighboring partitions and their lengths are updated accordingly. The indexes of $\mathcal{P}$ are then updated and the number of partitions $N$ is reduced by 1. This process continues until all partitions have a length of at least $\epsilon$.
            This partitioning result allows us to focus on individual semantic parts within the video and analyze their characteristics independently, enabling more detailed analysis and summary generation in subsequent stages.

    \subsection{Keyframes and Importance Scores}
    \label{subsec:method-keyframes-importance}
        After the partitioning step, the resulted partitions $\mathcal{P}$ are used to generate keyframes $\mathbf{k}$ which carry important information of the original input. An importance score ${v}_i$ is calculated for every sampled frame ${\hat{I}}_i$.
        
\begin{figure}[!t]
                \centering
                \includegraphics[width=0.5\paperwidth]{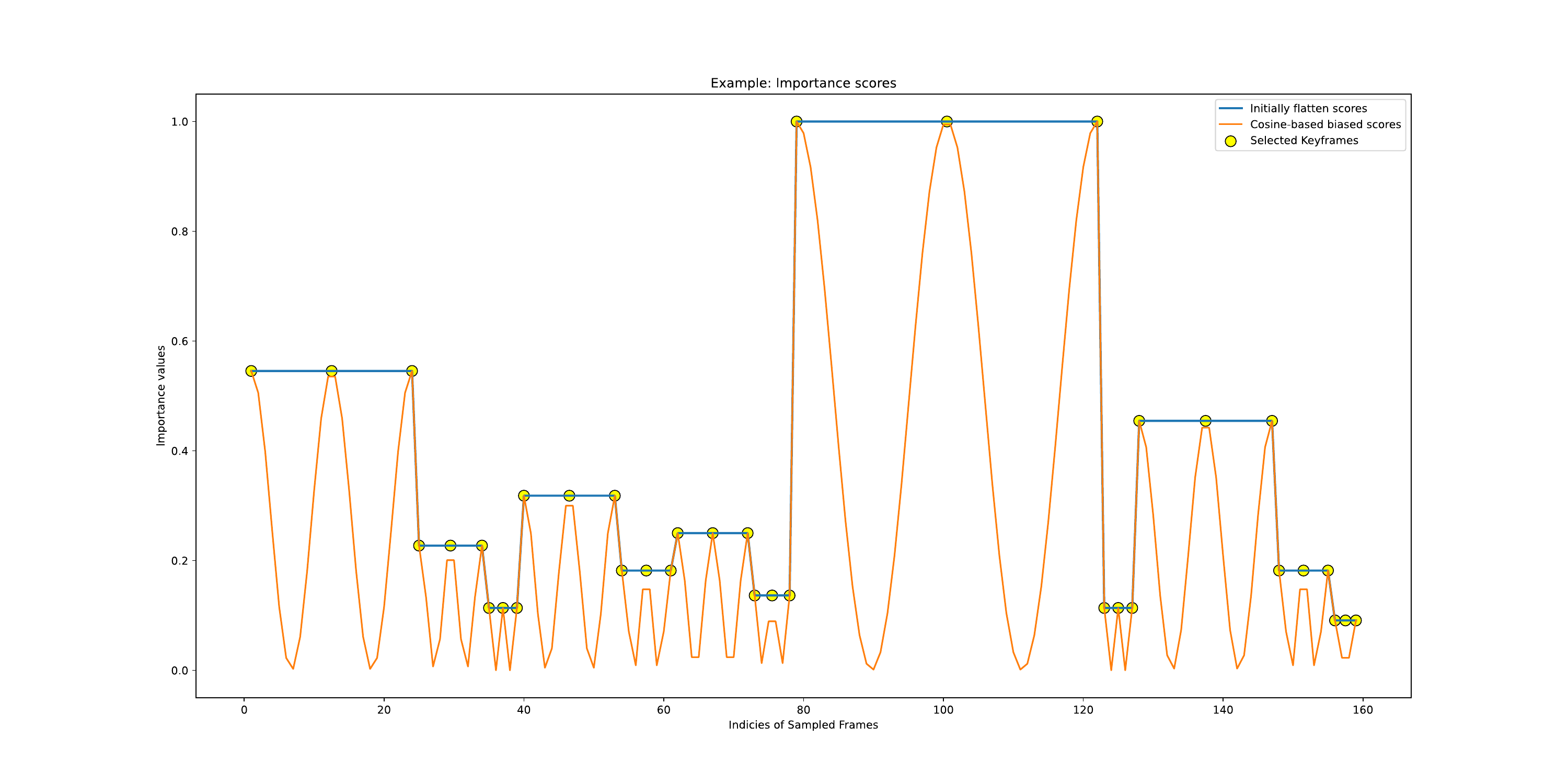}
                \caption{Comparison between cosine-interpolated scores and flat scores are demonstrated for two examples.}
                \label{fig:method-kf-imp}
            \end{figure}

        \paragraph{\textbf{Keyframes Selection}}
        \label{para:method-keyframes}
            The set of keyframes $\mathbf{k}$ is a subset of the indexes of sampled frames $\mathbf{k} \subset \mathbf{t}$, and is a union of partition-wise keyframes $\mathbf{k}^{(i)}$, that is $\mathbf{k} = \bigcup\limits_{i = 1}^{N} \mathbf{k}^{(i)}$.
            There are three options for extracting the partition-wise keyframes $\mathbf{k}^{(i)}$ from its associated partition $\mathcal{P}_i$ which are respectively \textit{Mean}, \textit{Middle}, and \textit{Ends}. These options can be further combined into more advanced settings such as the rule \textit{Middle} + \textit{Ends} demonstrated in Figure \ref{fig:method-kf-imp}.
            
        \paragraph{\textbf{Importance Scores}}
        \label{para:method-importance}
            The individual importance scores ${v}_i$ of all sampled frames ${\hat{I}}_i$ form a vector of importances $\mathbf{v} \in \mathbb{R}^{\hat{T}}$. We initialize the importance score $\mathbf{\hat{v}}$ to be the length of the section it belongs to. The final importance score of each sample ${v}_i$ is computed by scaling the initialized value ${\hat{v}}_i$ using a keyframe-biasing method.
            Several biasing options are given to either increase the importance of frames closer to keyframes or decrease the scores of others. Different interpolating methods are used to fill the importance scores of samples between key positions. Two options for interpolation are \verb|cosine| and \verb|linear|. An example illustrating the difference between cosine-interpolated importances and flat scores is given in Figure \ref{fig:method-kf-imp}. By determining the importance scores of frames, we can prioritize and select the most significant frames for inclusion in the video summary.

    \subsection{Implementation Details}
    \label{subsec:method-imple}

        Before the feature extraction step, the video is sampled with a target frame rate of $R = 4$ frames per second. We experiment with 2 pre-trained models to generate embeddings for each frame: DINO \cite{dino} and CLIP \cite{radford2021learning}. The input frame is processed using the pre-trained image processor associated with the pre-trained model. The output is an image, which is fed into the pre-trained model to obtain embeddings. For DINO, we select the first vector (cls token) in its output embedding as the semantic embedding of the sample. We concatenate the vector from all frames in to obtain the contextual embedding.

        For dimension reduction, we utilize models from \verb|scikit-learn|, including PCA and t-SNE. The number of clusters $K$ in contextual clustering is then computed by the equation provided in Section \ref{subsec:method-global-local}.
        In the semantic partition step, we set the window size $W$ for mode convolution to $5$, and the minimum length $\epsilon$ for each segment to $4$. For keyframe selection, we employ the setting \textit{Middle} + \textit{End}. In the importance scoring step, we use \verb|cosine| interpolation and set keyframe biasing scheme to \textit{Increase the importances of keyframes} with $B=0.5$.
        
    
\section{Experiments}
\label{sec:exp}

    
    \subsection{Dataset}
    \label{subsec:exp-data}
        
        For evaluating the performance of our TAC-SUM model, we utilize the SumMe dataset \cite{SumMe}. This benchmarking dataset consists of 25 videos ranging from 1 to 6 minutes in duration, covering various events captured from both first-person and third-person perspectives. Each video is annotated with multiple (15-18) key-fragments representing important segments. Additionally, a ground-truth summary in the form of frame-level importance scores (computed by averaging the key-fragment user summaries per frame) is provided for each video to support supervised training.
        
    \subsection{Evaluation Measures}
    \label{subsec:exp-eval}

        The summary selected by our summarizer is then compared with those generated by humans to determine its correctness, in other words, whether that summary is good or not depends on its similarity with regard to the annotated ones.
        A widely established metric for this comparison is f-measure, which is adopted in prior works \cite{apostolidis2020ac, apostolidis2021pglsum, chu2019spatiotemporal}. This metric requires an automatic summarizer to generate a proxy summary $\hat{\mathcal{S}}$ from pre-computed consecutive segmentations $\mathcal{S}$ associated with each video in the dataset. The f-measure metric is computed as f1-score between the segments chosen by automatic method against ground-truth selected by human evaluators. Previous studies \cite{apostolidis2021pglsum, zhang2016lstm} have formulated the conversion from importances to choice of segments as a Knapsack problem so that a simple dynamic programming method can be implemented to recover the proxy summary from outputted scores. The formulation includes lengths of segments as weighting condition while individual segment's value is computed using importance scores. More detailed information can be found in prior research \cite{zhang2016lstm}.


    \subsection{Comparison with State-of-the-art Methods}
    \label{subsec:exp-performance-compare}
        The performance of our proposed TAC-SUM approach is compared with various summarization methods from the literature in Table \ref{table:exp-compare}.
        These referenced approaches include both supervised and unsupervised algorithms that have been previously published, and the evaluation metric used is established under Section \ref{subsec:exp-eval}.
        As a general baseline, we include a random summarizer, which assigns importance scores to each frame based on a uniform distribution. The final performance is averaged over 100 sampling runs for each video \cite{apostolidis2020performance}.
    
        The results in Table \ref{table:exp-compare} highlight the effectiveness of our training-free approach, which achieves remarkable performance without any learning aspect. It outperforms existing unsupervised models by at least 3.18\%, demonstrating its ability to generate high-quality summaries. Moreover, our model ranks third when compared to state-of-the-art supervised methods, showing competitive performance and even surpassing several existing approaches.
    
        It is worth noting that the SMN method has been evaluated using only one randomly created split of the used data \cite{wang2019stacked}. Apostolidis \etal \cite{apostolidis2020performance} suggest that these random data splits show significantly varying levels of difficulty that affect the evaluation outcomes.
        
        We acknowledge that the pre-trained models used in our architecture were originally trained on general image datasets, which may not perfectly align with the distribution of the specific dataset used in this evaluation. Despite this potential distribution mismatch, our proposed method exhibits strong performance on the evaluated dataset, showcasing the generalizability and adaptability of this training-free framework.
            
        \begin{table}[!t]
        \centering
        \caption{Comparison of performance in f-measure (\%) among previous approaches and our method together with rankings on unsupervised only as well as in general.}
        \begin{tabular}{|cl|c|c|c|}
            \hline
            \multicolumn{2}{|c|}{\textbf{Methods}}                                                                                 & \textbf{F-Score}      & \begin{tabular}[c]{@{}c@{}}\textbf{Rank}\\ \textbf{(Unsupervised)}\end{tabular} & \begin{tabular}[c]{@{}c@{}}\textbf{Rank}\\ \textbf{(General)}\end{tabular} \\ \hline
            \multicolumn{2}{|c|}{Random summary}                                                                 & 40.2           & 7                                                             & 13                                                       \\ \hline
            \multicolumn{1}{|c|}{\multirow{5}{*}{Supervised}}   & SMN \cite{wang2019stacked}                     & \textbf{58.3}  & -                                                             & \textbf{1}                                               \\
            \multicolumn{1}{|c|}{}                              & VASNet \cite{fajtl2019summarizing}             & 49.7           & -                                                             & 10                                                        \\
            \multicolumn{1}{|c|}{}                              & PGL-SUM \cite{apostolidis2021pglsum}             & 57.1           & -                                                             & 2                                                       \\
            
            \multicolumn{1}{|c|}{}                              & H-MAN \cite{liu2019learning}                   & 51.8           & -                                                             & 5                                                        \\
            \multicolumn{1}{|c|}{}                              & SUM-GDA \cite{li2021exploring}                 & 52.8           & -                                                             & 4                                                        \\
            \multicolumn{1}{|c|}{}                              & SUM-DeepLab \cite{rochan2018sequence}          & 48.8           & -                                                             & 12                                                       \\ \hline
            \multicolumn{1}{|c|}{\multirow{6}{*}{Unsupervised}} & CSNet \cite{jung2019discriminative}            & 51.3           & 2                                                             & 6                                                        \\
            \multicolumn{1}{|c|}{}                              & AC-SUM-GAN \cite{apostolidis2020ac}            & 50.8           & 3                                                             & 7                                                        \\
            \multicolumn{1}{|c|}{}                              & CSNet+GL+RPE \cite{jung2020global}             & 50.2           & 4                                                             & 8                                                        \\
            \multicolumn{1}{|c|}{}                              & SUM-GAN-AAE \cite{apostolidis2020unsupervised} & 48.9           & 6                                                             & 11                                                       \\
            \multicolumn{1}{|c|}{}                              & SUM-GDA$_{unsup}$ \cite{li2021exploring}       & 50.0           & 5                                                             & 9                                                        \\
            \rowcolor{lightgray} \multicolumn{1}{|c|}{}                              & TAC-SUM (ours)                        & \textbf{54.48} & \textbf{1}                                                    & 3                                                        \\ \hline
        \end{tabular}
        \label{table:exp-compare}
        \end{table}
            
        
    \subsection{Ablation Study}
    \label{subsec:exp-ablation}

        To assess the contribution of each core component in our model, we conduct an ablation study, evaluating the following variants of the proposed architecture: variant \textbf{TAC-SUM w/o TC}: which is not aware of temporal context by skipping the semantic partitioning stage (Section \ref{subsec:method-global-local}), and the full algorithm \textbf{TAC-SUM (ours)}.
        The results presented in Table \ref{table:exp-ablation} demonstrate that removing the temporal context significantly impacts the summarization performance, thus confirming the effectiveness of our proposed techniques. The inclusion of temporal context enhances the quality of the generated summaries, supporting the superiority of our proposed TAC-SUM model.

        \begin{table}[!t]
        \caption{Ablation study based on the performance (F-Score(\%)) of two variants of the proposed approach on SumMe}
        \centering
            \begin{tabular}{|l|c|}
            \hline
            \multicolumn{1}{|c|}{\textbf{Settings}} & \textbf{F-Score} \\ \hline
            TAC-SUM w/o TC            & 46.00                   \\
            TAC-SUM (ours)           & \textbf{54.48}                 \\ \hline
            \end{tabular}
        \label{table:exp-ablation}
        \end{table}

        As mentioned in Section \ref{subsec:method-imple}, we conducted experiments using different pre-trained models for visual embedding extraction. Table \ref{table:exp-dino-clip} compares the result of the framework using different pre-trained models: dino-b16 and clip-base-16. Both models are base models with a patch size of 16. The "Best Config" column shows the configuration that achieved the best result, including the distance used in the clustering step (Euclidean), the algorithms used for embedding size reduction (PCA and t-SNE), and the dimension of the reduced embeddings represented by the number next to the reducer.
        The results presented in Table \ref{table:exp-dino-clip} demonstrate that our proposed framework performs relatively well with various pre-trained models, showcasing its flexibility and efficiency.
        The ability to work effectively with different pre-trained models indicates that our approach can leverage a wide range of visual embeddings, making it adaptable to various video summarization scenarios. This flexibility allows practitioners to choose the most suitable pre-trained model based on their specific requirements and available resources.
        
        \begin{table}[!t]
        \centering
        \caption{Comparison of performance (F-Score(\%)) with different embedding pre-trained models}
            \begin{tabular}{|cc|c|}
            \hline
            \multicolumn{2}{|c|}{\textbf{Setting}}                                       & \multirow{2}{*}{\textbf{F-Score}} \\ \cline{1-2}
            \multicolumn{1}{|c|}{\textbf{Embedding Model }} & \textbf{Best Config}       &                                   \\ \hline
            \multicolumn{1}{|c|}{dino-b16}                  & Euclidean PCA (34) + t-SNE (2) & \textbf{54.48}                    \\ \hline
            \multicolumn{1}{|c|}{clip-base-16}              & Euclidean PCA (44) + t-SNE (3) & 52.33                             \\ \hline
            \end{tabular}
        \label{table:exp-dino-clip}
        \end{table}

    \subsection{Qualitative Assessment}
    \label{subsec:exp-qualitative}

\begin{figure}[t!]
            \centering
            \includegraphics[width=\textwidth]{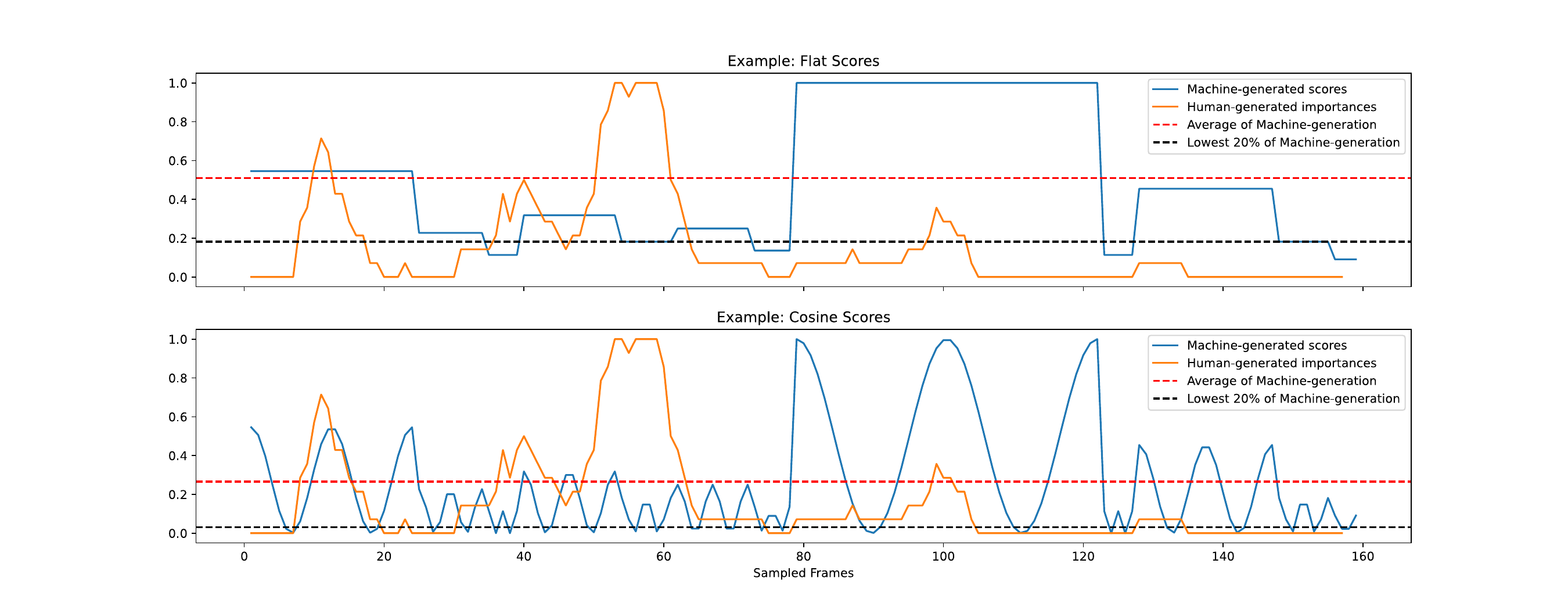}
            \caption{Comparison of importance scores between user-annotated scores and scores generated by the proposed method under the unbiased \texttt{flat} rule as well as the biased \texttt{cosine} rule.}
            \label{fig:exp-qualitative-score}
        \end{figure}

        To evaluate the interpretability of the proposed approach, we compared the automatically generated importance scores with those assigned by human annotators. Figure \ref{fig:exp-qualitative-score} displays the importance scores obtained through averaging human annotations as well as the scores generated by the proposed method.
        The flat result shows that each computed partition may be associated with one or several peaks in the user summaries, located at different positions within the partition. Longer partitions, which have higher flat scores according to the definition, tend to provide a more stable estimation of users’ peaks. The experimental result also provides insights into the keyframe-biasing method employed in the proposed method, wherein higher importance is assigned to frames that are closer to keyframes. This figure reveals that the majority of the peaks in the \verb|cosine| scores align with the peaks of the annotated importance. However, there are some peaks in the users' scores that are not captured by the \verb|cosine| interpolation.

\begin{figure}[t!]
            \centering
            \includegraphics[width=\textwidth]{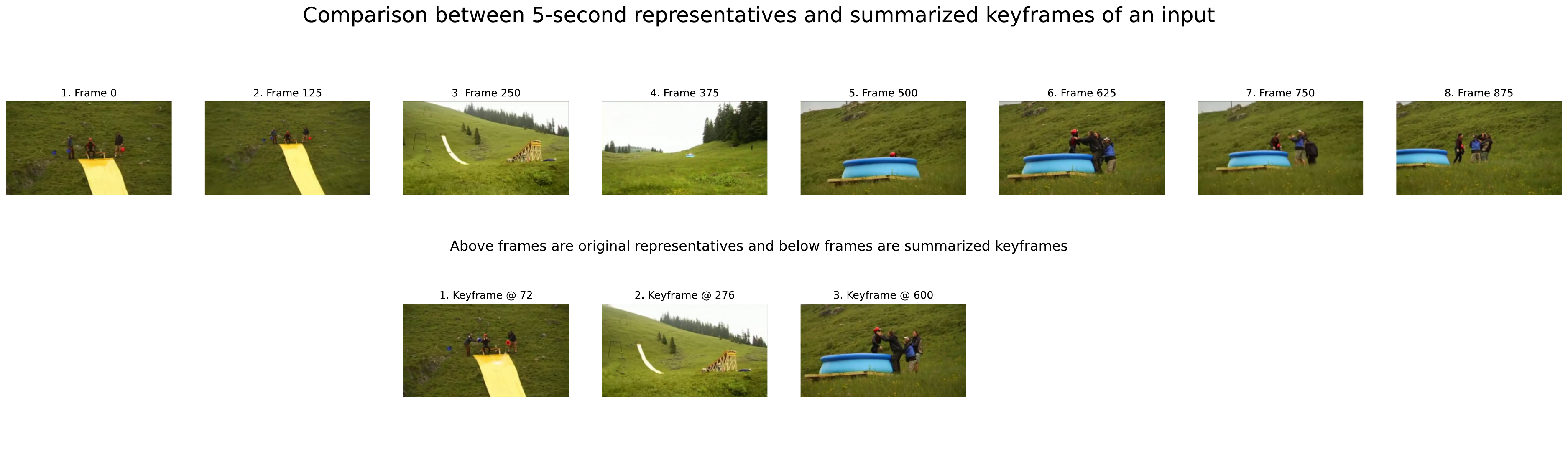}
            \caption{Comparison between the representatives sampled from the original video with its summarization as a set of keyframes.}
            \label{fig:exp-qualitative-video}
        \end{figure}

        A visual inspection of our method's summarization results is conducted in which a reference video is analyzed against its summary generated through the approach. We present the inspection's result in Figure \ref{fig:exp-qualitative-video} with the original frames of the reference video and selected keyframes. The original frames are sampled every 5 seconds from the video, which shows a man playing a game of sliding down a slope and jumping into a pool of water. The keyframes are selected based on their importance scores, which are higher than the average on the video level. Our method preserves the main content and events of the video and selects diverse and representative keyframes that show different aspects of the video. Our method generates informative and expressive keyframes that convey the main theme, message, or story of the video.
    
    \subsection{Limitations}
    \label{subsec:dis-disadvan}
        
        While the proposed method offers several advantages which have been already illustrated in the experimental results, it also has certain limitations that should be acknowledged. 
        \textbf{Naive rules} for scoring and selection of keyframes are being used. Therefore, our current approach may not always accurately predict frame importance. Incorporating more sophisticated scoring mechanisms can enhance the summarization process.
        \textbf{Limited learnability} is demonstrated by our method as it lacks the ability to improve in a data-driven way due to its current reliance on predefined rules. Future research could explore integrating data-driven approaches like machine learning algorithms or attention mechanisms to enhance adaptability.
    

\section{Conclusion}
\label{sec:conc}

    In this paper, we introduced TAC-SUM, an unsupervised video summarization approach that incorporates temporal context for generating concise and coherent summaries. The contextual clustering algorithm has successfully partitioned frames into meaningful segments, ensuring temporal coherence. Experimental results show that our method significantly outperforms traditional cluster-based approaches and even is competitive with state-of-the-art supervised methods on the SumMe dataset. 
    

    Despite its success, TAC-SUM has limitations related to pre-trained models and data-driven improvement. To address these limitations, future work will focus on integrating learnable components into the model to enhance the summarization process and improve adaptability to various video domains. This includes replacing the current algorithm for contextual clustering with a deep neural network having trainable parameters, enabling the model to capture more complex patterns and adapt to diverse video datasets. Additionally, various architectures and training techniques will be explored to transform the naive rules of importance into a data-driven scoring process, allowing complicated scores to be predicted.
    



\section*{Acknowledgement}
    This research is supported by research funding from Faculty of Information Technology, University of Science, Vietnam National University - Ho Chi Minh City.

\bibliographystyle{splncs04}
\bibliography{references}

\end{document}